\documentclass{article}

\usepackage{arxiv}

\usepackage[utf8]{inputenc} % allow utf-8 input
\usepackage[T1]{fontenc}    % use 8-bit T1 fonts
\usepackage{hyperref}       % hyperlinks
\usepackage{url}            % simple URL typesetting
\usepackage{booktabs}       % professional-quality tables
\usepackage{amsfonts}       % blackboard math symbols
\usepackage{nicefrac}       % compact symbols for 1/2, etc.
\usepackage{microtype}      % microtypography
\usepackage{lipsum}
\usepackage{graphicx}
\graphicspath{ {./images/} }

\usepackage{amssymb}
\usepackage{soul}
\usepackage{pdfcomment}  % Pour ajouter des descriptions accessibles aux images
\usepackage{float}
\usepackage{accessibility}
\usepackage{graphicx}    % Pour insérer des images
\usepackage{amsmath}
\title{District Vitality Index Using Machine Learning Methods for Urban Planners}

\author{
 Sylvain Marcoux \\
  Université du Québec à Trois-Rivières\\
  3351, boulevard des Forges \\
 Trois-Rivières (Québec), G8Z 4M3, Canada\\
  \texttt{sylvain.marcoux@cegeptr.qc.ca} \\
   \And
 Jean-Sébastien Dessureault \\
  Université du Québec à Trois-Rivières\\
  3351, boulevard des Forges \\
 Trois-Rivières (Québec), G8Z 4M3, Canada\\
  \texttt{jean-sebastien.dessureault@uqtr.ca} \\
 \\
}

\begin{document}
\maketitle

\begin{abstract}
City leaders face critical decisions regarding budget allocation and investment priorities. How can they identify which city districts require revitalization? To address this challenge, a Current Vitality Index (CVI) and a Long-Term Vitality Index (LVI) are proposed. These indexes are based on a carefully curated set of indicators. Missing data is handled using K-Nearest Neighbors (KNN) imputation, while Random Forest (RF) is employed to identify the most reliable and significant features. Additionally, k-means clustering is utilized to generate meaningful data groupings for enhanced monitoring of Long-Term Vitality. Current vitality is visualized through an interactive map, while Long-Term Vitality is tracked over 15 years with predictions made using Multilayer Perceptron (MLP) or Linear Regression (LR). The results, approved by urban planners, are already promising and helpful, with the potential for further improvement as more data becomes available. This paper proposes leveraging machine learning methods to optimize urban planning and enhance citizens' quality of life.
\end{abstract}

% keywords can be removed

\section{Introduction}
In a modern urban context, city districts play a central role as fundamental units of urban life. They are places where social bonds are formed, communities thrive, and the challenges and opportunities of urban life manifest. However, cities face numerous ongoing economic and social challenges, making it difficult to address all of these issues and make informed decisions. Cities face challenges such as rising rents and a low rental vacancy rate, which contribute to a housing crisis, as highlighted by \cite{nguyen_crise_2023}. Additionally, insufficient information on city indicators often leads to inequalities in resource allocation and investments between city districts, contributing to a decline in vitality in certain urban areas over time \footnote{\url{https://ici.radio-canada.ca/plusieurs_2019}}, regarding city revitalization projects. In this context, it becomes imperative to have a tool to monitor and evaluate the impact of these changes on the lives of citizens and city districts. Evaluating the vitality of urban districts is essential for better understanding local needs and improving the quality of life of urban residents.

The remainder of this paper is organized as follows: Section 2 details some research reviews, Section 3 presents the proposed methodology, Section 4 presents the results, Section 5 discusses the results and their implications, and Section 6 concludes the research.

\section{Literature Review} 

J. Jacobs, author of \textit{The Death and Life of Great American Cities} \cite{jacobs_death_1993}, was among the first to focus on urban vitality, highlighting elements such as urban density, land use diversity, and urban planning. Jacobs criticizes modern city planning for large-scale developments that prioritize structural expansion over the needs and scale of human interaction. In her methodology, she prioritizes empirical observation over large theoretical models. Also, J. E. Drewes and M. Van Aswegen \cite{drewes_determining_2010}, and X. Li and al. \cite{li_data_2020}, define vitality as the ability of an entity to survive or function effectively. Thus, an urban Vitality Index represents the capacity of an urban center to remain viable, meet community needs, and enhance residents' quality of life. Drewes and Van Aswegen’s index is constructed using normative scores for welfare, satisfaction, social dynamics, and spatial distribution, calculated based on the average quartiles of the associated variables. Meanwhile, Dessureault et al. \cite{dessureault_unsupervised_2021} include eight variables, such as material and social deprivation indices, housing condition, property values, vacancy rates, renovation permit values, and rents. They use a genetic algorithm and k-means to cluster 135 areas into 10 clusters, while also using RF for weighting features. On the other hand, Haynes, Hook, Chiodi Grensing, and Ecklund \cite{haynes_analysis_2018} propose a Metropolitan Vitality Index (MVI) with 24 indicators, including the number of small businesses and physical activity spaces. The studied areas were ranked from 1 to 381 for each indicator, and those rankings were then summed. Finally, the MVI values were recalculated within a range of 80 to 120, with higher scores indicating better performance. K. Scott \cite{scott_katherine_katherine_2010} suggests a Community Vitality Index based on the direction of data trends over time for variables such as property crimes, volunteerism, and violent crimes. Analogously, Liu, Gou, and Xiong \cite{liu_vital_2022} examine growth, diversity, and mobility indicators, such as youth proportion, economic and racial diversity, and service proximity. MinMax standardization is applied to the indicators, and feature weighting is performed using the entropy method, along with scores provided by urban planning scholars. Additionally, correlations between different dimensions of urban vitality are analyzed. Herath and Mittal \cite{herath_adoption_2022} provide a comprehensive review of smart cities and artificial intelligence (AI), citing Singapore and Zurich are among the top ten smart cities. Eighty percent of studies on urbanization and machine learning use clustering, as detailed by Wang and Biljecki \cite{wang_unsupervised_2022}, with popular algorithms such as k-means, self-organizing maps, and DBSCAN. Unsupervised algorithms are also used, as demonstrated by Dessureault \cite{dessureault_unsupervised_2021} and Li \cite{li_data_2020}. To evaluate the effectiveness of clustering, P.J. Rousseeuw introduced the Silhouette Score \cite{rousseeuw_silhouettes_1987}, a metric that measures how well points are grouped within their respective clusters. The k-means algorithm, used for the classification of areas, was originally developed by S. Lloyd \cite{lloyd_least_1982}. This unsupervised clustering method is renowned for its simplicity, efficiency, and speed, as it minimizes the sum of squared distances between data points and their respective cluster centers, called 'centroids'. Additionally, data imputation is performed using K-Nearest Neighbors, introduced by Thomas Cover and Peter Hart in 1967 \cite{cover_nearest_1967}. Furthermore, feature weighting is done using the RF algorithm, introduced by Leo Breiman in 2001 \cite{breiman_random_2001}.

\section{Methodology}

\begin{figure} [H] %[!htbp]
    \centering
      \pdftooltip{\includegraphics[width=1\linewidth]{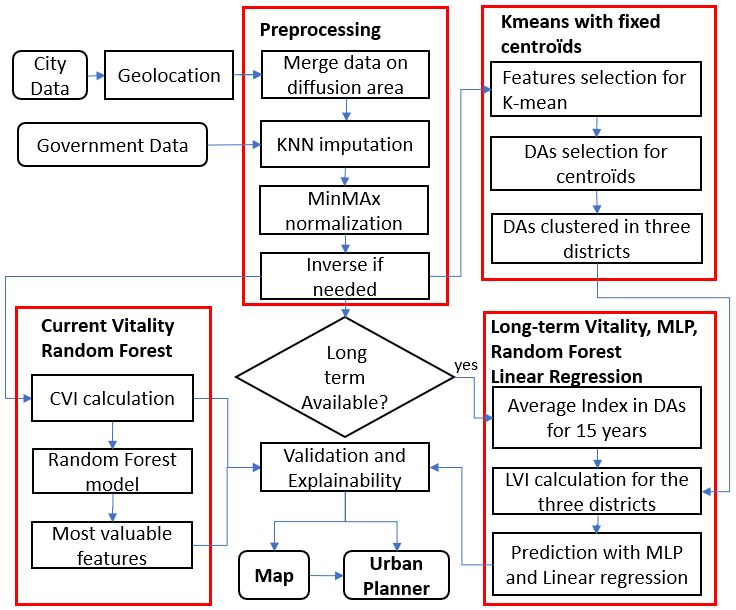}}{Figure 1 presents the architecture of the methodology. The preprocessing box represents the preprocessing stage, which includes merging data, KNN imputation, Min-Max normalization, and inverse transformation of the data if needed. The second box represents the K-means clustering of DAs into three districts, including feature selection, selecting DAs for centroids, and clustering with K-means. The CVI box corresponds to the CVI calculation, highlighting the most important features using Random Forest (RF). The LVI box represents the LVI calculation for the three districts over 15 years and the 2026 prediction using either MLP or LR. The center of Figure \ref{diagram} shows the explainability stage and the outputs of the process, including maps and valuable insights for urban planners.}
    \caption{Method's architecture}
    % \Description{Figure 1 presents the architecture of the methodology. The preprocessing box represents the preprocessing stage, which includes merging data, KNN imputation, Min-Max normalization, and inverse transformation of the data if needed. The second box represents the K-means clustering of DAs into three districts, including feature selection, selecting DAs for centroids, and clustering with K-means. The CVI box corresponds to the CVI calculation, highlighting the most important features using Random Forest (RF). The LVI box represents the LVI calculation for the three districts over 15 years and the 2026 prediction using either MLP or LR. The center of Figure \ref{diagram} shows the explainability stage and the outputs of the process, including maps and valuable insights for urban planners.}
    \label{diagram}
\end{figure}

In this paper, two indices are developed. On the one hand, the CVI allows us to assess the vitality of city districts at the present moment using 13 variables. On the other hand, the LVI allows us to monitor four variables over 15 years. Figure \ref{diagram} illustrates the architecture of the framework. The preprocessing box represents the preprocessing stage, including imputation, normalization, and inverse transformation of the data. The second box represents the Kmeans clustering of DAs into three districts. The CVI box corresponds to the CVI calculation, highlighting the most important features using RF. The LVI box represents the LVI calculation for the three districts and the 2026 prediction using either MLP or LR. The center of Figure \ref{diagram} presents the explainability stage and the outputs of the process, including maps and valuable insights for urban planners.

\subsection{Current Vitality Index}
The CVI is made up of 13 variables that come from two main sources: the Government of Canada and the City of Shawinigan. Data are first obtained in groupings called dissemination areas (DAs). A dissemination area (DA) is a small and relatively stable geographic unit composed of one or more adjacent dissemination blocks. DAs cover all the territory of Canada.  These groups (DAs) can be visualized in Figure \ref{HeatMap}. The indicators included in the Index are: \textit{Proportion of apartments that need minor and major repairs}, \textit{Material and Social Deprivation Index},\textit{ Average housing costs for owners}, \textit{Average renter housing costs}, \textit{Unoccupied housing rate}, \textit{Average value of renovations}, \textit{Average value of construction}, \textit{Average value of buildings}, \textit{Number of vacant buildings}, \textit{Proportion of young people}, \textit{Number of businesses}. A Google Geolocation API has been utilized to obtain \textit{Number of businesses}. Additionally, for \textit{The average cost of renovation and construction}, this API was essential for linking addresses to their corresponding dissemination areas. For data preprocessing, a MinMax scaling function was applied. The specific preprocessing steps for each feature are detailed in Table \ref{table:1}.

\begin{table}[!htbp]
    \centering
    \begin{tabular}{|l|l|}
        \hline
        \textbf{Treatments} & \textbf{Indicators} \\ 
        \hline
        MinMax & All of them \\ 
        \hline
        Inversion & Material and Social deprivation Index \\ 
                  & Unoccupied housing rate \\ 
                  & Number of vacant buildings \\ 
                  & Prop. of apartments that need minor repairs\\
                  & Prop. of apartments that need major repairs \\
        \hline
        Imputation & Material and Social deprivation Index \\ 
                   & Average tenant housing costs \\ 
        \hline
    \end{tabular}
    \vspace{0.2cm}
    \caption{Feature's preprocessing}
    \label{table:1}
\end{table}

The CVI for dissemination area $k$ is calculated with (\ref{Index})
\begin{equation}
    CVI_k = \frac{1}{n} \sum_{i=1}^{n} x_i
    \label{Index}
\end{equation}
Where $CVI_k$ is the CVI for the dissemination area $k$, $n$ is the number of variables, $x_i$ is the value of variable $i$ for DA $k$ and $k$ represents the DA. In this study, the calculation of the CVI includes 13 variables ($n$) across 87 dissemination areas ($k$). To visualize the boundaries of dissemination areas (DAs) on a map of the city based on their CVI, a Python procedure was developed using the \texttt{Geopandas} library. This interactive map allows users to visualize the values of indicators for a selected dissemination area. Additionally, the DAs are color-coded (heat map) based on the CVI values (Figure \ref{HeatMap}).  Another procedure has also been implemented to visualize the boundaries of DAs with low indices compared to those with high indices.

\subsection{Variable Importance in Vitality Index}

A critical question is determining which variable has the greatest influence on the Current Vitality Index. Identifying this variable would enable city policymakers to strategically invest in areas where improvements in this feature could enhance vitality scores. The RF algorithm was used to rank variables by importance. RF is a supervised learning algorithm based on decision trees. Each optimal split within a tree reduces data impurity, measured by the Gini index. The reduction in impurity is calculated by comparing impurity before and after the split. The target variable for this algorithm, the CVI, was selected. The \texttt{RandomForestRegressor} implementation from the \textit{scikit-learn} library was used with the parameter \texttt{n\_estimators=500}, wich represents the number of decision trees. The variables were ranked by importance and visually presented in a bar chart to communicate the results to urban planners. To validate the results obtained with \texttt{RandomForestRegressor}, the \texttt{XGBRegressor} algorithm from the \texttt{xgboost} library was also employed to assess variable importance. This algorithm, renowned for its efficiency, speed, and performance in handling large datasets, is widely recognized in predictive modeling competitions, as highlighted by T. Chen and C. Guestrin in their work, "XGBoost: A Scalable Tree Boosting System" \cite{chen_xgboost_2016}.

\subsection{Long-Term Vitality Index}

To derive an exploitable LVI and enable an in-depth analysis, it was necessary to group the 87 dissemination areas. The \texttt{K-means} unsupervised algorithm from the \textit{scikit-learn} library was used for this clustering. While initial cluster centers are typically chosen randomly, alternative methods exist for their selection, as highlighted in the article \textit{FC-KMeans: Fixed-centred k-means algorithm} \cite{ay_fc-kmeans_2023}. In this study, the \texttt{init} parameter was used to define these initial centroids. Specifically, the chosen centroids correspond to the characteristics of three districts: Urban, Residential, and Commercial. This classification facilitates the comparison and interpretation of trends across several years within these different types of districts. In agreement with urban planners, the following variables were used to optimize the clustering procedure: \textit{Population density per square kilometer}, \textit{Proportion of young people}, \textit{Number of businesses}, and \textit{Average building value}. This dimensionality reduction from the initial 13 variables ensured coherent clustering and a strong Silhouette score. The selected variables were chosen, in accordance with urban planner, to clearly differentiate the three defined districts. The \textit{Population density} indicator characterizes urban city districts, while the \textit{Proportion of young people} helps distinguish residential areas. Finally, the \textit{Number of businesses} and \textit{Average building value} variables facilitate clustering dissemination areas within the commercial sector. For the selection of initial centroids, a dissemination area that most accurately represents each of the three groups was chosen. To assess the cohesion and quality of the clustering, the Silhouette Index was used. The dissemination areas within the three districts were used to calculate the LVI for these groupings. An average of all indicators within each sector was computed to derive three Long-Term Vitality Indices, one for each sector, for the following years: {2006, 2011, 2016, 2021}. A MLP type neural network, with two hidden layers, and LR are used to perform regression for the next census year, 2026. The \texttt{MLPRegressor} and \texttt{LinearRegression} algorithms from the scikit-learn library are employed for this task. To evaluate the quality of the predictions, the least squares method is applied, and the mean\_squared\_error function from the sklearn.metrics library is used to calculate the error.

The LVI for sector $s$ is calculated with (\ref{IndiceLT}) 
\begin{equation}
    LVI_s = \frac{1}{nbrAD_s} \sum_{i=1}^{nbrAD_s} CVI_i
    \label{IndiceLT}
\end{equation}  
Where $LVI_s$ is the LVI for sector $s$, $nbrAD_s$ is the number of dissemination areas within sector $s$, $CVI_i$ is the CVI of dissemination area $i$ and \\ $s \in \{\text{Urban, Commercial, Residential}\}$

\subsection{Explainability}
\begin{figure} [!htbp]
    \centering
    \pdftooltip{\includegraphics[width=0.8\linewidth]{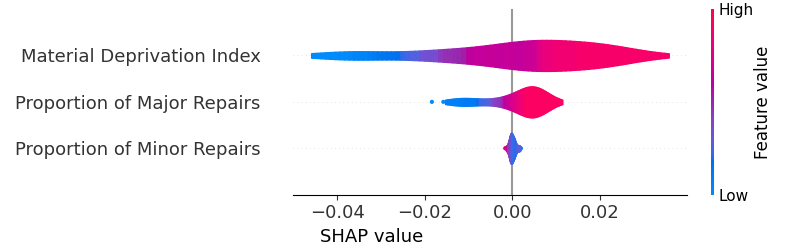}}{This figure presents an example of SHAP values displayed using a violin plot.}
    \caption{SHAP impact on model output magnitude}
    \label{SHAP value RF}
    % \Description{This figure presents an example of SHAP values displayed using a violin plot.}
\end{figure}

To further explain the results of the \texttt{RandomForestRegressor} regarding the most influential variables in the CVI, SHAP (Shapley Additive Explanations) values were computed and represented using violin plots, as shown, for example, in Figure \ref{SHAP value RF}. In this figure, the positive red SHAP values indicate that the feature is positively correlated with higher CVI values in the regression model's predictions \cite{louhichi_shapley_2023}. SHAP is a post-hoc method that explains models through an independent, model-agnostic process external to the model itself. In contrast, ante-hoc methods are inherently interpretable, meaning their interpretability is intrinsic to the model's structure and decision-making process. The clustering of dissemination areas used in the LVI will also be validated and explained using SHAP values. Additionally, radar charts will be generated to assess the internal consistency of each grouping (Figures \ref{radar_kmeans}). An interpretation of all these values will be automatically generated through a Python procedure and included in a PDF file. This file will feature both explanatory charts and interpretative text. The charts will include violin plots and horizontal bar plots for SHAP values. A distinct chart will be generated for each defined sector: Urban, Residential, and Commercial. Using these tools, city decision-makers will gain a clearer understanding of the factors influencing the ranking generated by the RF algorithm and the clustering of dissemination areas with k-means. This analysis will help them identify key variables impacting the rankings and enable more informed decision-making based on these insights.

\section{Results}

\subsection{Current Vitality Index}

\begin{figure} [H] %[!htbp]
    \centering 
    \pdftooltip{\includegraphics[width=0.7\linewidth]{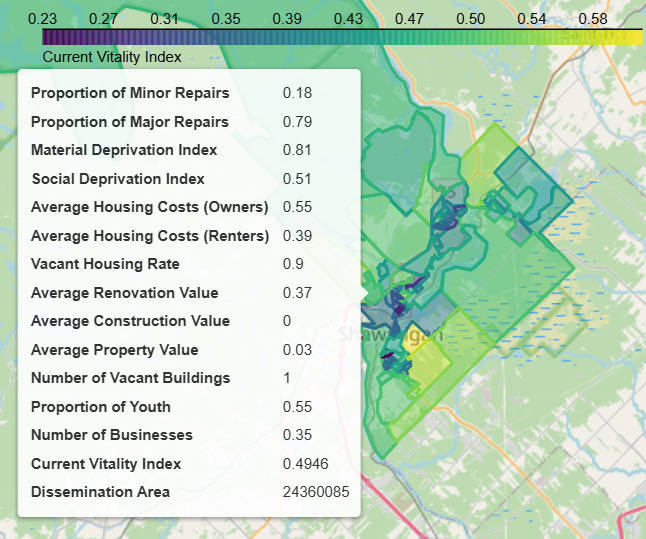}}{The heatmap displays the CVI for a given DA along with the values of all its features. Additionally, DAs are color-coded based on their CVI: darker shades indicate a lower CVI, while lighter shades represent a higher CVI.}
    \caption{Heat Map of Current Vitality Index}
    \label{HeatMap}
    % \Description{The heatmap displays the CVI for a given DA along with the values of all its features. Additionally, DAs are color-coded based on their CVI: darker shades indicate a lower CVI, while lighter shades represent a higher CVI.}
\end{figure}

To analyze the distribution of the CVI, a histogram was created by dividing the data into eight classes. The resulting histogram indicates that the CVI follows a normal distribution. Figure \ref{HeatMap} represents an interactive heat map based on the CVI. Users can visualize the values of the indicators by selecting a dissemination area. In Figure \ref{HeatMap}, dissemination area 24360085 is selected.

% enlevé pour 5 pages
% \begin{figure} [htbp]
%     \centering
%     \includegraphics[width=0.7\linewidth]{histogram.png}
%     \caption{Histogram}
%     \label{Histogram}
% \end{figure}

%\pdftooltip{\includegraphics[width=0.8\textwidth]{image.png}}{Schéma montrant l'impact des variables sur le CVI.}

\subsection{Long-Term Vitality Index}

\begin{figure}  [!htbp]
    \centering
    \pdftooltip{\includegraphics[width=.9\linewidth]{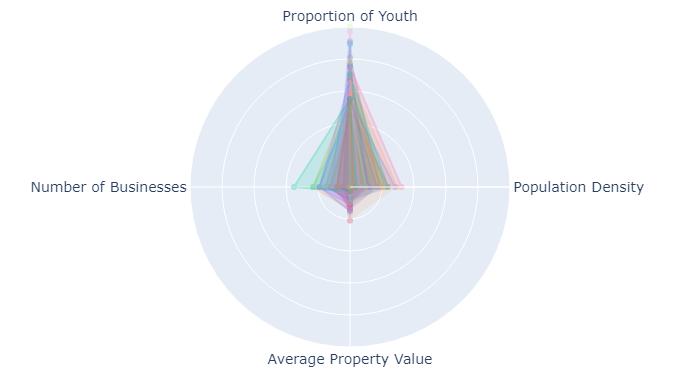}}{This figure presents the radar chart for the Residential cluster. This cluster exhibits a high degree of internal consistency due to its strong alignment across the key features, as the data points within this cluster share similar characteristics.}
    \caption{Residential cluster of K-means}
    \label{radar_kmeans}
    % \Description{This figure presents the radar chart for the Residential cluster. This cluster exhibits a high degree of internal consistency due to its strong alignment across the key features, as the data points within this cluster share similar characteristics.}
    
\end{figure}

The clustering of dissemination areas using the K-means algorithm with the three fixed centroids for the LVI can be analyzed through stacked radar charts. Figure \ref{radar_kmeans} presents the radar chart for the \textit{Residential} cluster. A visual analysis reveals that this cluster exhibits a high degree of internal consistency. In contrast, the \textit{Commercial} cluster consists of dissemination areas with more varied characteristics. With this information, urban planners can validate the three clusters.

\begin{figure}  [!htbp]
    
    \centering  
    \pdftooltip{\includegraphics[width=.9\linewidth]{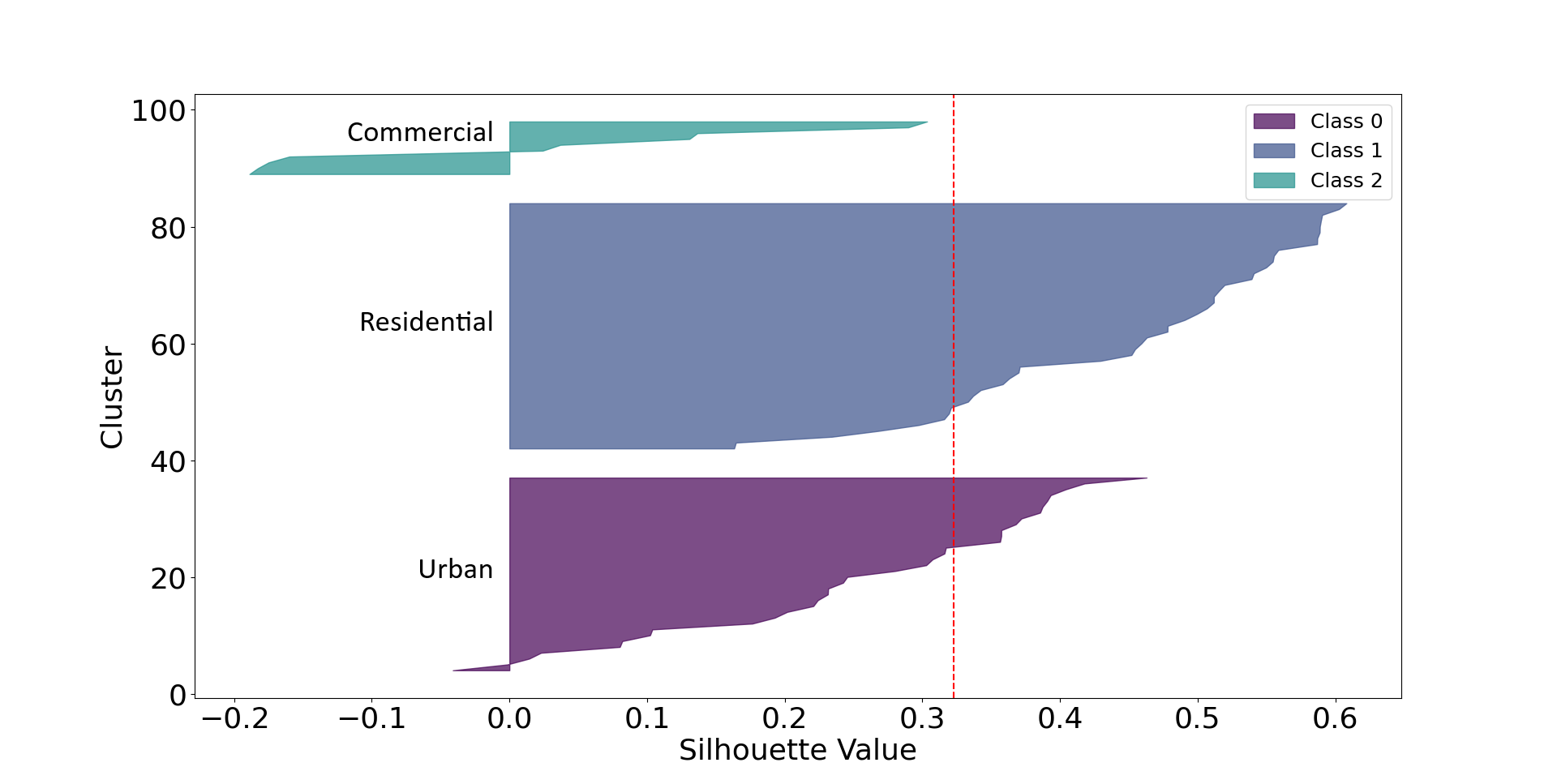}}{This figure presents the clustering consistency assessment using the Silhouette score for the three clusters. Cluster 1 has a higher score, indicating a well-defined cluster, while Cluster 2 has a lower or negative score, suggesting potential misclassification or poor clustering.}
    \caption{Clustering consistency using Silhouette score}
    \label{Silhouette score}
    % \Description{This figure presents the clustering consistency assessment using the Silhouette score for the three clusters. Cluster 1 has a higher score, indicating a well-defined cluster, while Cluster 2 has a lower or negative score, suggesting potential misclassification or poor clustering.}
\end{figure}

As shown in Figure \ref{Silhouette score}, the average Silhouette score obtained by the K-means algorithm is 0.3229 as shown by the dotted red line. While this index is not exceptionally high or close to 1, it indicates moderate cohesion between points within the clusters and partial separation between the clusters. The three colors represent the clusters, and the varying bandwidths indicate that some clusters include more data than others. Additionally, negative values in the first cluster suggest that those points may be misclassified. 

\begin{figure}  [!htbp]  
    \centering
    \pdftooltip{\includegraphics[width=1\linewidth]{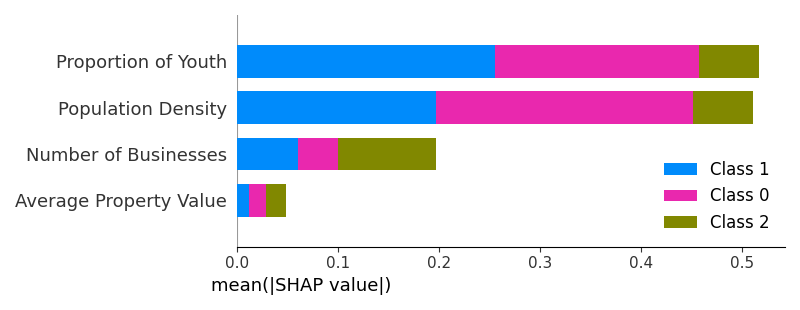}}{  
SHAP values for Class 0 (Urban):
Mean SHAP values per feature:
Population Density        0.254441
Proportion of Youth       0.201608
Number of Businesses      0.039313
Average Property Value    0.017124
SHAP values for Class 1 (Residential):
Mean SHAP values per feature:
Proportion of Youth       0.255756
Population Density        0.196712
Number of Businesses      0.060992
Average Property Value    0.012123
SHAP values for Class 2 (Commercial):
Mean SHAP values per feature:
Number of Businesses      0.097023
Population Density        0.058856
Proportion of Youth       0.058834
Average Property Value    0.019227
}
    \caption{SHAP Average impact on model output magnitude}
    \label{SHAP value}
%     \Description{SHAP values for Class 0 (Urban):
% Mean SHAP values per feature:
% Population Density        0.254441
% Proportion of Youth       0.201608
% Number of Businesses      0.039313
% Average Property Value    0.017124
% SHAP values for Class 1 (Residential):
% Mean SHAP values per feature:
% Proportion of Youth       0.255756
% Population Density        0.196712
% Number of Businesses      0.060992
% Average Property Value    0.012123
% SHAP values for Class 2 (Commercial):
% Mean SHAP values per feature:
% Number of Businesses      0.097023
% Population Density        0.058856
% Proportion of Youth       0.058834
% Average Property Value    0.019227}
\end{figure}

SHAP values are used to improve the explainability of the k-means clustering results. In Figure \ref{SHAP value}, the average SHAP values show that the variables \textit{Proportion of Youth} and \textit{Population Density} are the most influential in the Urban (Class 0) and Residential (Class 1) categories. In contrast, the variable \textit{Number of Businesses} plays a key role in defining the Commercial (Class 2) category. Finally, the \textit{Average Property Value} indicator has minimal influence on the overall clustering. 

\begin{figure} [!htbp]
    \centering
    \pdftooltip{\includegraphics[width=1\linewidth]{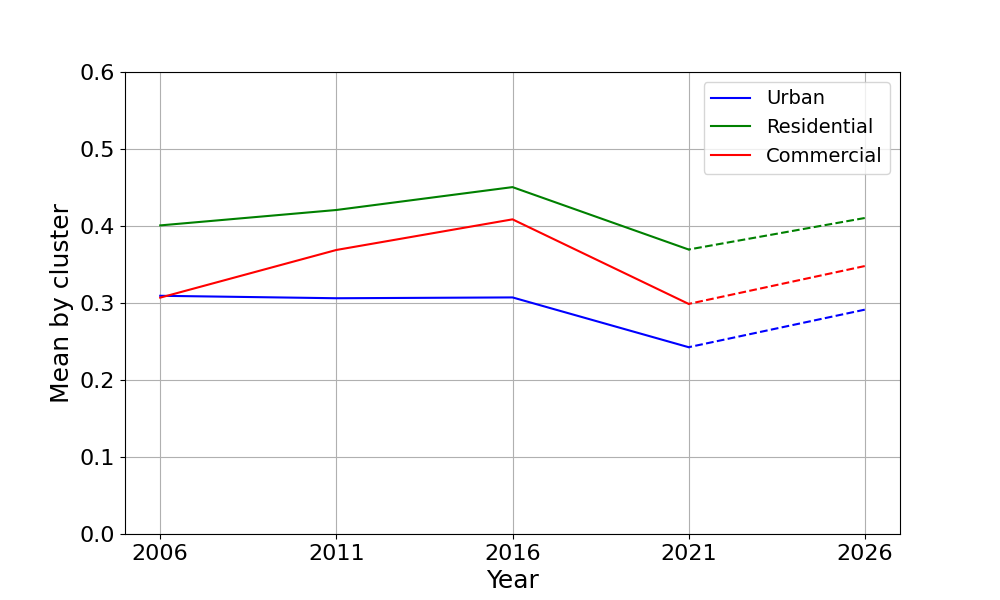}}{This figure shows the time series of the Long-Term Vitality Index (LVI) for different clusters. 
    Values for the Urban cluster are:(2006, 0.31), (2011, 0.31), (2016, 0.31), (2021, 0.24), (2026, 0.29)
    Values for the Residential cluster are:(2006, 0.40), (2011, 0.42), (2016, 0.45), (2021, 0.37), (2026, 0.41)
    Values for the Commercial cluster are:(2006, 0.31), (2011, 0.37), (2016, 0.41), (2021, 0.30), (2026, 0.35)}
    \caption{Time series of the LVI}
    \label{time series}
    %  \Description{This figure shows the time series of the Long-Term Vitality Index (LVI) for different clusters. 
    % Values for the Urban cluster are:(2006, 0.31), (2011, 0.31), (2016, 0.31), (2021, 0.24), (2026, 0.29)
    % Values for the Residential cluster are:(2006, 0.40), (2011, 0.42), (2016, 0.45), (2021, 0.37), (2026, 0.41)
    % Values for the Commercial cluster are:(2006, 0.31), (2011, 0.37), (2016, 0.41), (2021, 0.30), (2026, 0.35)}
\end{figure}

Figure \ref{time series} illustrates the evolution of the LVI from 2006 to 2021, segmented by the three neighborhood clusters: Urban, Residential, and Commercial. The figure also includes projections for 2026, shown as dotted lines, which are obtained using a LR model. Table \ref{tab:EQM} presents the Mean Squared Errors (MSE) for each predictive model. Among the three models evaluated, LR emerges as the most accurate, achieving the lowest MSE.

%  enlevé pour 5 pages  
\begin{table} [!htbp]
    \centering
    \small
    \begin{tabular}{|c|c|c|c|}
        \hline
        \textbf{Cluster} & \textbf{MLP Regressor} & \textbf{RF} & \textbf{LR} \\
        \hline
        Urban & 0.0022 & 0.0017 & 1.3e-33 \\
        \hline
        Residential & 0.0023 & 0.00052 & 2.8e-32 \\
        \hline
        Commercial & 0.0023 & 0.00036 & 0.00017 \\
        \hline
    \end{tabular}
    \caption{Mean Squared Errors of the Models by Cluster}
    \label{tab:EQM}
    \vspace{0.2cm}
\end{table}

\section{Discussion}
It can be challenging to obtain very high values when we have more than two dimensions in our data. This justifies the dimensionality reduction to 3, rather than using all of our features. However, the obtained clusters align well with the three types of neighborhoods defined for the city: Urban, Residential, and Commercial. This suggests that the initial centroids chosen for the clustering were relevant for this classification. Furthermore, urban planners in Shawinigan have confirmed that the dissemination areas within the clusters correspond closely to the city's realities. These findings indicate that the three clusters can effectively illustrate long-term trends. Analyzing the clustering using SHAP values highlights the significant impact of demographic characteristics on the classification of dissemination areas. However, the Commercial cluster remains more challenging to delineate compared to the others. A general decline in the LVI is observed in all three clusters between 2016 and 2021. This decrease may be related to the global COVID-19 pandemic, which impacted the clusters' various economic and social aspects, reducing their vitality. The LR results suggest a potential stabilization or possible recovery by 2026, although the trends remain uncertain and depend on future economic and social conditions. This combined approach, which takes advantage of both historical data and AI-based predictions, provides a comprehensive overview of the past and future evolution of neighborhood vitality in Shawinigan. Table \ref{tab:improvement} highlights the enhancements made over the Vitality Index (VI) of Dessureault et al. \cite{dessureault_unsupervised_2021}.  Urban planners utilizing the two Vitality Indexes introduced here can analyze urban vitality more effectively using the interactive maps. Furthermore, the clustering approach, which employs fixed centroids, results in significant clusters validated by urban planners.
\begin{table}[!htbp]
    \centering
    \scriptsize
    \renewcommand{\arraystretch}{1.2} % Adjust row height for better readability
    \setlength{\tabcolsep}{6pt} % Adjust column spacing
    \begin{tabular}{|c|c|c|}
        \hline
        \textbf{Aspects} & \textbf{Dessureault Vitality Index (VI)} & \textbf{CVI and LVI} \\ 
        
        \hline
        Imputation with RF & X & \checkmark \\
        \hline
        Identification of Significant Clusters & X & \checkmark \\
        \hline
        Control Over Regression Models & X & \checkmark \\
        \hline
        Control Over Indicators & X & \checkmark \\
        \hline
        Interactive Map Features & X & \checkmark \\
        \hline
    \end{tabular}
    \caption{Comparison of improvements over Dessureault et al. \cite{dessureault_unsupervised_2021}}
    \label{tab:improvement}
   % \vspace{0.2cm}
\end{table}

\section{Conclusion}
This paper examines the vitality of Shawinigan's city districts using two key indicators: the CVI and the LVI. By classifying 87 dissemination areas into Urban, Residential, and Commercial categories through k-means clustering, significant socio-economic dynamics were identified. These indexes are designed to monitor vitality and identify priority areas for investment within the city. In summary, our study found that residential areas exhibited greater vitality, supported by favorable socioeconomic conditions and active investments in renovations and construction. In contrast, urban and commercial zones require targeted interventions to enhance quality of life and drive revitalization.  These findings underscore the effectiveness of monitoring vitality through these indexes, which serve as essential tools for achieving this paper objective. We encountered limitations in data accessibility due to the smaller size of the city studied. Unlike larger cities in Canada, smaller ones often lack comprehensive and readily available data, which posed challenges for this analysis. In addition, extensive research on the indicators available over a 15-year period could be conducted to improve the LVI. This research provides a solid foundation for intelligent and sustainable urban planning, aiming to significantly improve citizens' quality of life.

%\section{Appendices}
\bibliographystyle{ACM-Reference-Format}
\bibliography{District_Vitality}

\appendix


\end{document}